\documentclass{article}
\usepackage{spconf,amsmath,graphicx,hyperref}

\usepackage{cite}
\usepackage{amssymb,amsfonts}
\usepackage{algorithmic}
\usepackage{textcomp}
\usepackage{xcolor}
\usepackage{mathtools}
\usepackage[font=small,labelfont=bf, figurename=Figure, tablename=Table]{caption} 
\usepackage{subcaption}
\usepackage{wrapfig}

\expandafter\def\expandafter\normalsize\expandafter{%
    \normalsize%
    \setlength\abovedisplayskip{2pt}%
    \setlength\belowdisplayskip{8pt}%
    \setlength\abovedisplayshortskip{-4pt}%
    \setlength\belowdisplayshortskip{2pt}%
}

\DeclareMathSymbol{\shortminus}{\mathbin}{AMSa}{"39}
\begin{document}

\title{Feature Identification for Hierarchical Contrastive Learning}

\name{Julius Ott$^{1,3}$, Nastassia Vysotskaya$^{2,3}$, Huawei Sun$^{1,3}$, Lorenzo Servadei$^{1}$, Robert Wille$^{1}$}

\address{
$^{1}$Technical University Munich, 
$^{2}$Friedrich-Alexander-Universität Erlangen, Nürnberg \\
$^{3}$Infineon Technologies AG, Neubiberg }
\maketitle

\begin{abstract}
Hierarchical classification is a crucial task in many applications, where objects are organized into multiple levels of categories. However, conventional classification approaches often neglect inherent inter-class relationships at different hierarchy levels, thus missing important supervisory signals. Thus, we propose two novel hierarchical contrastive learning (HMLC) methods. The first, leverages a Gaussian Mixture Model (G-HMLC) and the second uses an attention mechanism to capture hierarchy-specific features (A-HMLC), imitating human processing. Our approach explicitly models inter-class relationships and imbalanced class distribution at higher hierarchy levels, enabling fine-grained clustering across all hierarchy levels. On the competitive CIFAR100 and ModelNet40 datasets, our method achieves state-of-the-art performance in linear evaluation, outperforming existing hierarchical contrastive learning methods by $2$ percentage points in terms of accuracy. The effectiveness of our approach is backed by both quantitative and qualitative results, highlighting its potential for applications in computer vision and beyond.
\end{abstract}
\section{Introduction}\label{sec:intro}
\begin{figure}[ht]
    \centering
    \includegraphics[width=0.8\linewidth]{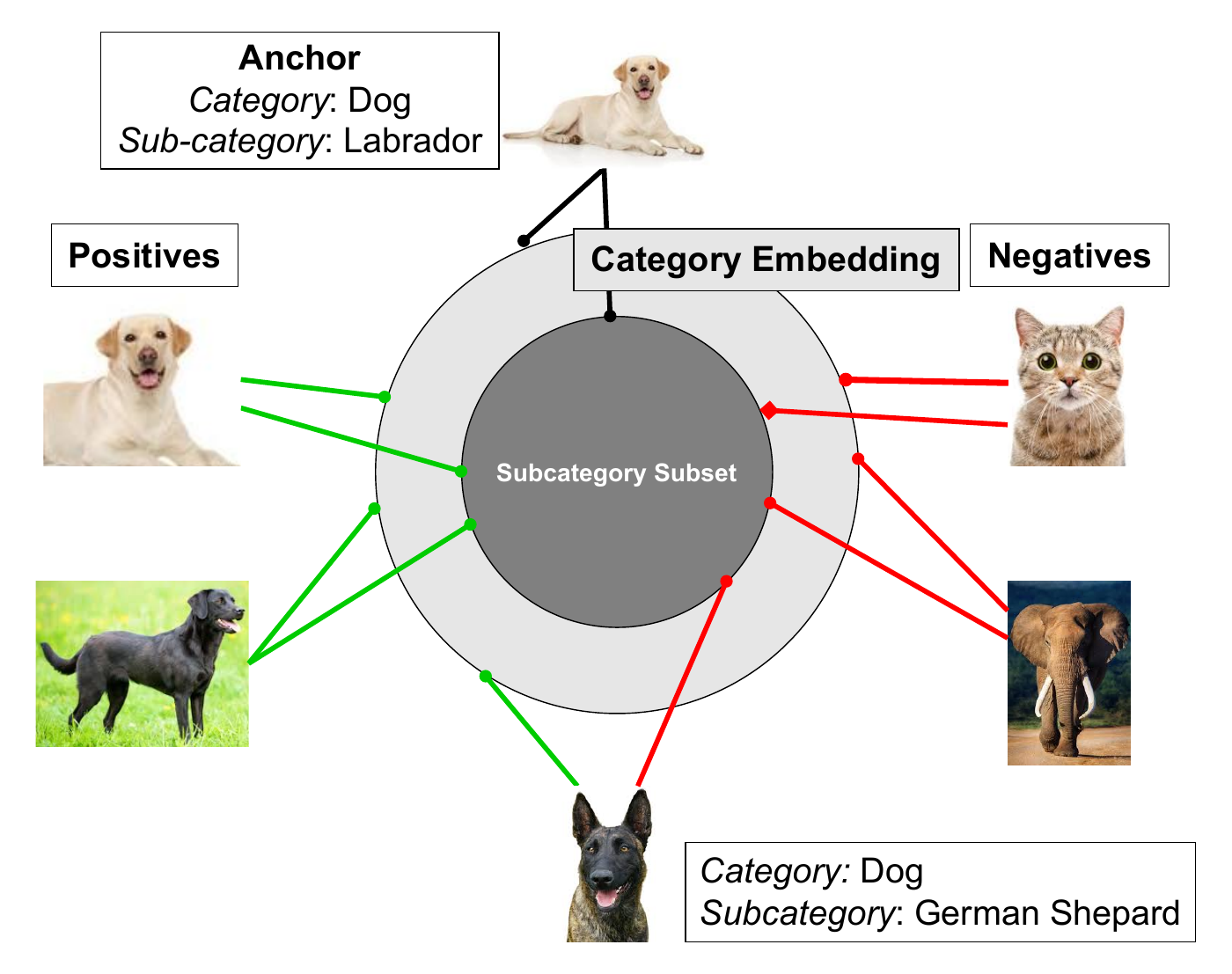}
    \caption{\textit{Hierarchical multi-label contrastive learning (HMLC) setup. The dogs are positive pairs in the first level but negative pairs in the second level. Whereas the cat and elephant are negatives on both levels.}}
    \label{fig:gmlc}
\end{figure}

\noindent With the rise of AI solutions in our daily life, classification remains a prominent application across various domains, such as vision~\cite{yolo}, natural language processing~\cite{text_class}, or discriminating generated images~\cite{gan}.
In the domain of machine learning-based classification, the conventional approach to learning has been to organize classes into a flat list. However, in real-world applications, hierarchical multi-labeling occurs naturally and frequently, as exemplified by its presence in biological classification (see Fig. \ref{fig:gmlc}), e-commerce product categorization, and retail spaces. The hierarchical representation serves to efficiently capture relationships between different classes, yet this valuable information is often underutilized in learning tasks.
In representation learning frameworks, a single embedding function must generalize to unseen downstream tasks and data. Thus, this embedding function must represent the data concisely and accurately, including the preservation of the hierarchical representation in the embedding space. 

In recent years, several unsupervised~\cite {simclr, selfproperties, caron2020unsupervised} and supervised metric learning~\cite{supcon, lar, triplet} frameworks have been proposed that rely on minimizing the distance between representations of positive pairs and maximizing the distance between negative pairs. 
However, these approaches frequently prove inadequate in supporting multi-label learning and leveraging information about the inter-label relationships. In this context, hierarchical multi-label contrastive learning (HMLC) methods have been proposed~\cite{hcsc, hmce}. HMLC methods augment multiple labels to single objects and impose constraints on the hierarchy of these class levels.
%In MulCon~\cite{mulcon}, an individual network generates embeddings of the input image under the context of a specific label. The label-level embeddings of the same class are then trained with a contrastive loss. Further, the label embeddings can be used as multivariate prior in the Gaussian mixture variational autencoder (C-GMVAE)~\cite{bai2022gaussian}.
%The approach from~\cite{MLC} improves the definition of positive samples based on their intersection with related images, considering cases where two images have one or all objects in common. 
To achieve hierarchical clustering, it is crucial to preserve the hierarchical structure of the labels in the embedding vector, where each level of the hierarchy is represented by a subset of features. The embedding vector encodes a hierarchical representation of the categories, with each level implicitly defined by a subset of features. \newline
\indent To enforce this hierarchical coarse-to-fine clustering, HCSC~\cite{hcsc} learns unsupervised representations based on the highest hierarchy level. These embeddings are then clustered to select semantically unrelated negative samples.
HMCE~\cite{hmce} is a successor that leverages supervised contrastive loss (SupCon)~\cite{supcon} for each level of the hierarchy, operating on one vector for all hierarchies. 
In turn, the fidelity of the subcategories varies significantly, with some categories having a rich hierarchical structure (e.g., "dog`` can be divided into sub-categories like "golden retriever`` and "poodle``) while others are flatter (e.g., "wolf`` has no sub-categories). This variability in fidelity highlights the need for a mechanism to identify the relevant features for each level of the hierarchy, rather than relying on a one-for-all approach. Unlike previous approaches that apply contrastive learning uniformly across all levels of the hierarchy, our method recognizes the importance of identifying relevant features for each level, and proposes a feature identification mechanism to capture nuanced relationships between categories and subcategories.
In conclusion, our proposed method represents a significant advancement in hierarchical clustering, offering a robust and flexible framework for identifying sub‐level features and achieving superior cluster performance and linear evaluation. Its ability to handle disbalanced classes and complex hierarchical structures makes it a promising solution for a wide range of applications. Further, its superiority over state‐of‐the‐art approaches is clearly demonstrated by our experimental results.
\begin{figure*}[h!]
    \centering
    \includegraphics[width=0.8\linewidth]{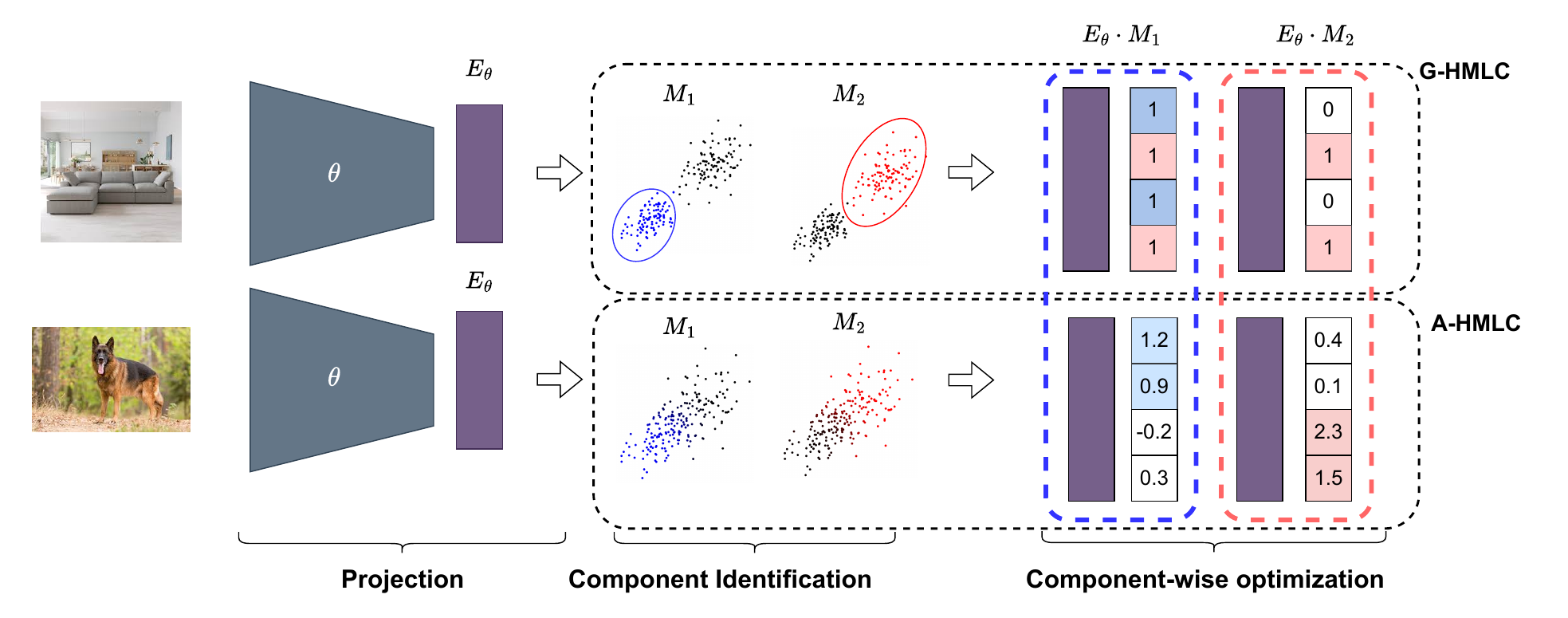}
    \caption{\textit{Illustration of the G-HMLC and A-HMLC architectures.
    The projection head $E_{\theta}$ maps the images to embedding vectors. In G-HMLC, a GMM is fitted on this embedding vector to generate a mask for each hierarchy level. This hard masking is suitable for unrelated lower hierarchy classes (furniture-couch $\neq$ furniture-table). For shared features along the hierarchy tree (e.g. dogs), A-HMLC computes soft attention scores. The hierarchical binary or soft masks, both denoted as $M_{i}$ for readability, are then multiplied by the feature vector.}}
    \label{fig:gmlc_structure}
\end{figure*}
\vspace{-10pt}
\section{Related Work}
\label{sec:related_work}
\noindent This section first provides an overview of contrastive learning methods. Then, hierarchical multi-label contrastive learning applications and corresponding losses are presented.
\vspace{-10pt}
\subsection{Contrastive Learning}
\noindent Contrastive learning has received a lot of attention as a pre-training method for self-supervised learning for images~\cite{chopra2005learning, simclr}, text~\cite{clip}, audio~\cite{baevski2020wav2vec}, and sequential data~\cite{tcn}. It can be summarized as \textit{learning by comparison}. For this purpose, a data sample is chosen as an anchor. A positive sample is then of the same class, while a negative sample is from a different, "contrasting`` class. The learning objective is minimizing the distance between anchor and positive sample, and maximizing it between anchor and negative sample.
The key ingredients for improvements in contrastive methods are augmentations~\cite{simclr, autoaugment} and batch size. Augmentations should significantly change the visual appearance while preserving semantic information. Large batch sizes are typically used, but memory issues have led researchers to investigate memory banks to access previous representations~\cite{moco}. \newline
\indent Supervised contrastive learning addresses the challenge of negative mining by using label information. The SupCon loss~\cite{supcon} is a generalized approach that uses multiple positives and negatives. Given a neural network encoder $E_{\theta}(x)$, the SupCon loss minimizes the distance between embeddings of the current sample $z_i$ and those with the same label $z_p$ and maximizes the distance to negative embeddings $z_a$, given a temperature $\tau$.
\begin{equation}\label{eq:supcon}
    \mathcal{L}^{SupCon} = \sum_{i \in \mathcal{I}} \frac{\shortminus1}{|P(i)|} \sum_{p \in P(i)} \log \frac{exp(\mathbf{z}_i \cdot \mathbf{z}_p/\tau)}{\sum_{a \in A(i)} exp(\mathbf{z}_i \cdot \mathbf{z}_a / \tau)},
\end{equation}

However, real-world scenarios often have multiple objects per image, making a single label per image suboptimal.
\subsection{Hierarchical Multi-Label Contrastive Learning}
\noindent Hierarchical Multi-Label Contrastive Learning (HMLC) focuses on multiple labels per object in a hierarchical order. In a hierarchical dataset, each object has multiple labels $y_i^h$ where $h \in {1,..., H}$ defines the hierarchical order. This additional information enriches the label space and defines inter-label correlations. \newline
\indent Previous work introduced a hierarchical cost function as a regularizer to enforce hierarchical ordering~\cite{guided}. The HMC loss~\cite{hmce} leverages the SupCon loss in the hierarchical framework, defining positive and negative pairs for each level in the hierarchy, denoted as $z_p^h$ and $z_a^h$ respectively. The HMC loss is defined as:
\begin{align}\label{eq:hmc}
    \mathcal{L}^{HMC} &= \sum_{h = H}^{1}\frac{1}{|H|} \sum_{i \in \mathcal{I}} \frac{\shortminus\lambda_h}{P^h(i)} \sum_{p \in P^h(i)} L^{pair}(i, h) \\
  \mathcal{L}^{pair}(i, h) &= \log \frac{exp(\mathbf{z}_i \cdot \mathbf{z}^h_p/\tau)}{\sum_{a \in A(i,h)} exp(\mathbf{z}_i \cdot \mathbf{z}^h_a / \tau)},
\end{align}
where $\lambda_h = f(h)$ controls the penalty for each level in the hierarchy.
The HMC loss is extended to the Hierarchical Multi-label Contrastive Enforcing (HMCE) loss, which optimizes solely with respect to the largest loss across all hierarchies. The HMCE loss is defined as:
\begin{equation}\label{eq:hmce}
    \begin{split}
        &\mathcal{L}^{HMCE} =  \\
        &\frac{1}{|H|} \sum_{h=H}^{1} \sum_{i \in \mathcal{I}} \frac{-\lambda_h}{P^h(i)} \sum_{\mathclap{p \in P^h(i)}} \max \left\{ \mathcal{L}^{pair}(i,h),\mathcal{L}_{max}^{pair}(i,h\shortminus1) \right\}
    \end{split}
\end{equation}
\vspace{-14pt}
\section{Methodology}
\begin{figure}
  \centering
    \includegraphics[width=0.8\linewidth]{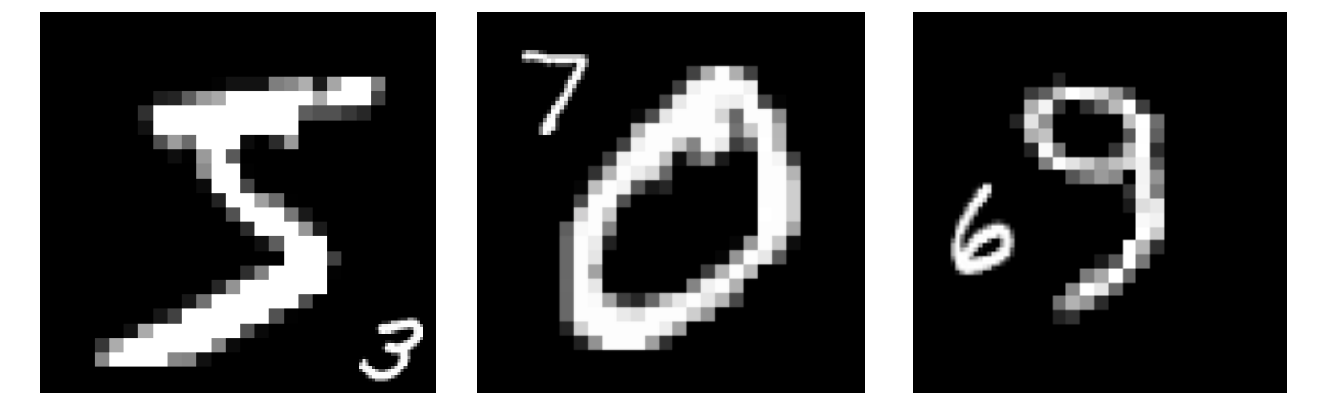}
  \caption{\textit{Examples of the hierarchical MNIST dataset. The central digit refers to the class, and the image size is scaled from $32\times32$ to $192\times192$ pixels. The subsidiary digit around denotes the category and is placed randomly with a size of $32\times32$.}}
  \label{fig:mnist_example}
\end{figure}

\noindent In this paper, we propose feature-based hierarchical multi-label
contrastive learning methods. Intuitively, it is reasonable to perform contrastive learning only on that part of the embedding vector that is relevant for a certain hierarchy level. To this end, we propose two ways to identify the relevant features, which are detailed in Figure \ref{fig:gmlc_structure}. First, we apply a Gaussian Mixture Model (GMM), which masks relevant features in the embedding before contrastive learning (G-HMLC) is performed. 
To this end, two embedding vectors are generated: the anchor image and the same image with random augmentations. These vectors define a two-dimensional plane, where a two-dimensional GMM is fitted with prior knowledge about the number of hierarchy levels. The GMM then predicts the masks $M_h$ for each hierarchy level. 
These masks are incorporated into Eq. \ref{eq:hmce}, where we replace $\mathbf{z}_i$ with $\mathbf{z}_i[M_h]$, $\mathbf{z}_a$ with $\mathbf{z}_a[M_h]$ and $\mathbf{z}_p^h$ with $\mathbf{z}_p^h[M_h]$ and define it as $\mathcal{L}^{pair}(i,h)[M_h]$.
Then the SupCon loss is applied to the masked feature vectors for each level. The GMM is applied during training for each image in the batch individually, and is initialized with the previously identified mean values. This setup optimizes the convergence speed, which further reduces the number of iterations. %Additionally, we decrease the GMM iterations such that the obtained masks are noisy at the beginning, but they converge as training progresses. In order to facilitate implicit regularization, multiple GMMs are employed in parallel. This approach improves the efficiency of the process and mitigates overfitting during mask generation.
Intuitively, this hard masking performs well when the lower hierarchy samples are clearly distinguishable, for instance, in \ref{fig:mnist_example}, or when we aim to distinguish different furniture, it is clear that a wardrobe is significantly different from a chair. \newline 
However, some data points share features across the hierarchy tree. To address this, we propose the second approach: attention-based multi-label hierarchical contrastive loss (A-HMLC), using soft masking via multiplicative attention maps. Following the self-attention implementation of \cite{vaswani2017attention}, we define $K$, $Q$ as single-layer neural networks, which take the embedding vector as input. The attention weights are defined as:
\begin{equation}
    attention = softmax\left(\frac{Q \cdot K}{\sqrt{d}}\right),
\end{equation}
where $d$ is the dimension of the embedding vector. The attention weights are multiplied by the embedding vector before the contrastive loss is applied. In addition, we employ one attention head per hierarchy level, such that each head learns the relevant features for the specific level. I both approaches, a linear scaling was employed compared to the exponential scaling in the HMCE loss.
\vspace{-8pt}
\section{Experiments}\label{sec:experiments}
\noindent The subsequent experiments provide a comprehensive evaluation of the Gaussian (G-HMLC) and attention-based (A-HMLC) feature extraction methods, assessing their performance through both qualitative and quantitative analyses.
\begin{figure}[ht!]
\centering
\begin{subfigure}{.2\textwidth}
    \includegraphics[width=\textwidth]{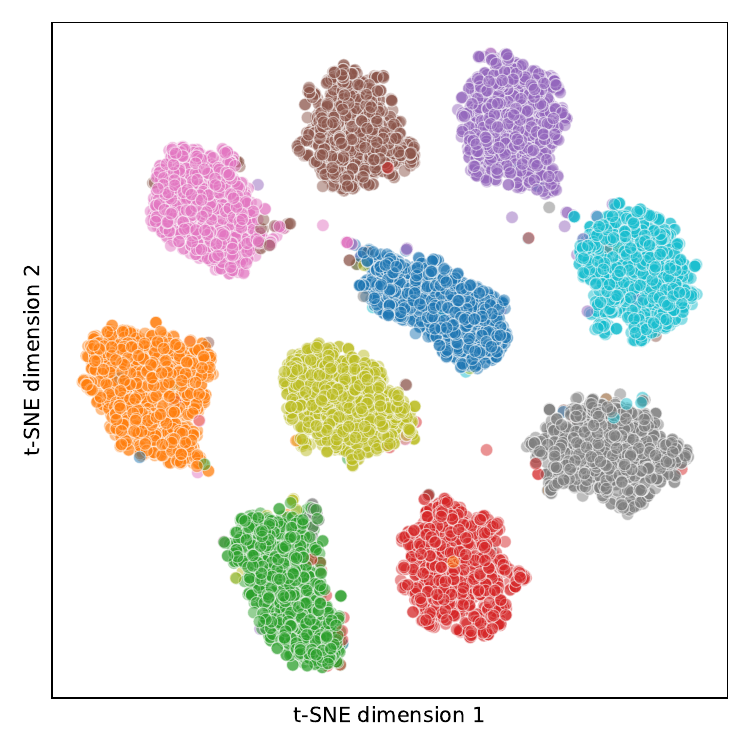}
\end{subfigure}
\begin{subfigure}{.2\textwidth}
    %\centering
    \includegraphics[width=\textwidth]{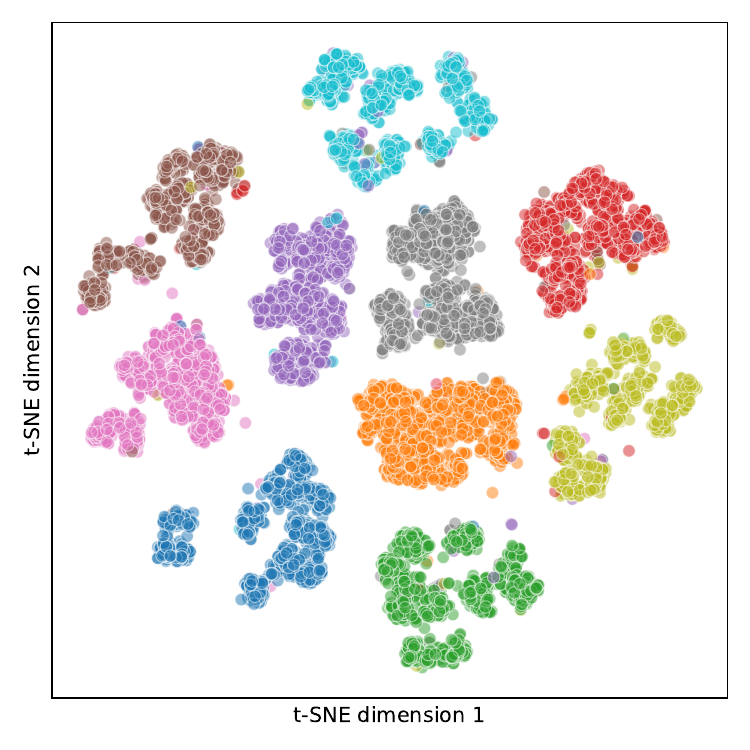}
\end{subfigure}%
\vfill
\begin{subfigure}{0.2\textwidth}
    \includegraphics[width=\textwidth]{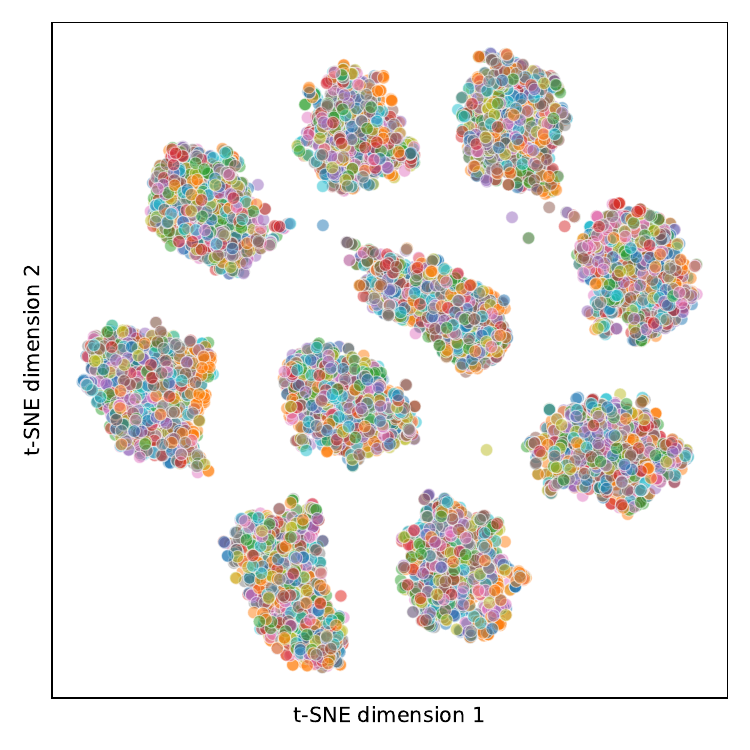}
    \caption{HMCE}
\end{subfigure}
\begin{subfigure}{0.2\textwidth}
    \includegraphics[width=\textwidth]{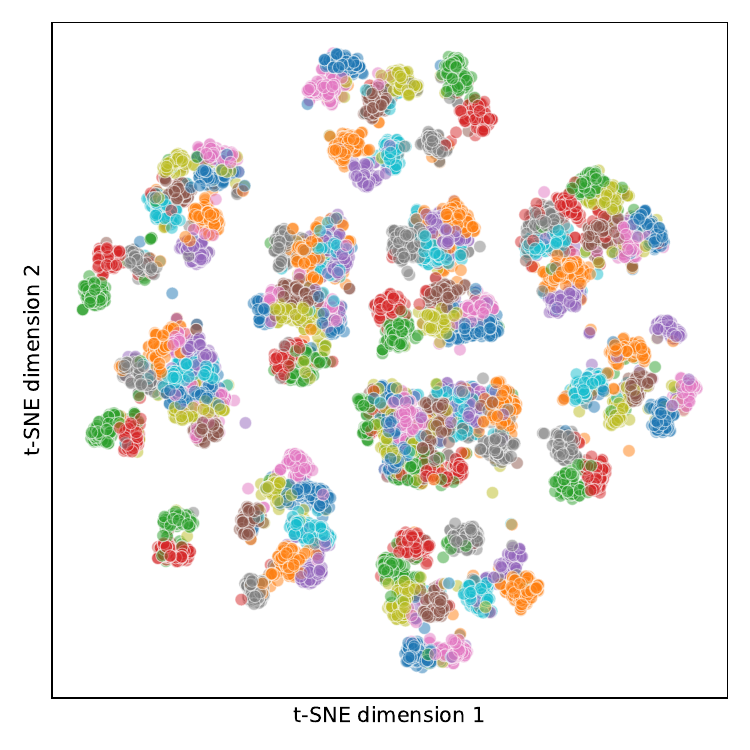}
    \caption{G-HMLC (ours)}
\end{subfigure}
\caption[short]{\small \textit{First and second t-SNE components of the hierarchical MNIST test embeddings. The HMCE loss (left column) separates the first level (top row) but misses a clear separation for the second level (bottom row). The proposed G-HMLC loss (right column) separates both levels.}}
\label{fig:mnist_experiments}
\end{figure}
\subsection{Qualitative Evaluation}
\noindent The experiments under consideration encompass the state-of-the-art hierarchical contrastive HMCE loss with our proposed G-HMLC loss. 
In this experiment, the objective is to evaluate the clustering quality of the hierarchical contrastive losses for each hierarchy level.
To this end, a hierarchical MNIST dataset is created, as shown in Figure \ref{fig:mnist_example}.
Note that the category and sub-category are distinct, making the G-HMLC the appropriate choice here.
In Figure \ref{fig:mnist_experiments} the first and second t-SNE~\cite{tsne} components of the test dataset embeddings are presented. The HMCE loss demonstrates effective separation at the first level but exhibits a lack of distinction at the second level. This observation underscores the significance of the ordering of the hierarchy in the HMCE loss formulation, and a noteworthy performance is anticipated when the class is at the first level and the categories represent the higher-order categories. The proposed G-HMLC loss attains a remarkable first- and second-level separation. 
At the same time, $3\%$ more variance in the embeddings are expressed by the G-HMLC loss, underlining the more effective dimensionality reduction.
\subsection{Quantitative Evaluation}
\noindent The experiments under consideration encompass the hierarchical contrastive losses. As a baseline, we utilize the single-label contrastive SupCon loss and non-contrastive cross-entropy loss. 
To assess its efficacy for more complex tasks and diverse data modalities, we examined the balanced CIFAR100~\cite{cifar} image dataset and the unbalanced ModelNet40~\cite{modelnet} point-cloud dataset. The CIFAR100 dataset comprises 100 classes, each originating from 20 distinct categories, where each category has 5 sub-classes. On the contrary, the ModelNet40 dataset is grouped into 6 categories with the number of sub-categories ranging from 2 to 26.  We conduct a ResNet50~\cite{resnet} as an encoder for CIFAR100 and a PointNet++ encoder~\cite{qi2017pointnet++} for ModelNet40.
The test accuracy values presented in Table \ref{tab:results} have been obtained in accordance with the linear evaluation scheme outlined in~\cite{simclr}.
The proposed A-HMLC and G-HMLC losses improve the performance on the balanced CIFAR100 dataset. As assumed in the introduction, the attention-based A-HMLC loss further outperforms the G-HMLC loss since many classes share features across hierarchy levels. In addition, all hierarchical contrastive variants outperform the single-label losses.
In contrast, the HMCE loss falls short against the single-label losses using the unbalanced ModelNet40 dataset. This highlights the aforementioned issue on the fidelity of the sub-classes, which is not addressed by the HMCE loss. Since the proposed methods update each feature independently, they are more robust to the unbalanced data. 
\subsection{Hierarchical Ordering}
\noindent To assess the impact of hierarchy ordering, an ablation study was conducted, where the ordering of the labels is reversed, which impacts the algorithmic behavior due to hierarchical scaling. In the presence of a perfect hierarchical clustering, the order ought to be irrelevant. %For the HMCE and feature-based HMC losses, the reverse ordering imposes more relevance on the category due to the hierarchy scaling. Although the hierarchy scaling enforces the hierarchical correctness, it also reduces the importance of features from higher-order categories. Thus, reversing the order can illustrate whether the loss considers all hierarchical features or mainly focuses on a single level. 
The numerical evaluations, presented in Table \ref{tab:ablation}, demonstrate that the G-HMLC and A-HMLC exhibit enhanced robustness compared to the HMCE loss. This quantitative finding is corroborated by the clustering quality exhibited in Figure \ref{fig:mnist_experiments}. Consequently, it can be deduced that feature-based HMC losses possess the capacity to learn fine-grained hierarchical clusters that are resilient to label order variations.
\begin{table}
    \centering
    \begin{tabular}{l|cc}
        Method &  ModelNet40 & CIFAR100\\ \hline
        Cross Entropy & 90.5 & 75.3 \\
        SupCon & 91.2 & 75.26\\
        \hline
        HMCE &  90.2 & 75.95 \\
        G-HMLC (ours)  & $\mathbf{91.5}$ & $\mathbf{76.13}$ \\
        A-HMLC (ours) & $\mathbf{92.42}$ & $\mathbf{76.19}$ \\
    \end{tabular}
    \caption{\textit{Test accuracy after linear evaluation on ModelNet40 and CIFAR100 datasets.}}
    \label{tab:results}
\end{table}
\begin{table}
    \centering
    \begin{tabular}{l|cc}
        Method &  ModelNet40 & CIFAR100\\ \hline
        HMCE category first & 85.4 & 73.21\\
        HMCE class first &  90.2 & 75.95 \\
        \hline
        G-HMLC category first & 91.05 & 74.36\\
        G-HMLC class first& $\mathbf{91.5}$ & $\mathbf{76.13}$ \\
        A-HMLC category first& $\mathbf{92.38}$ & $75.69$ \\
        A-HMLC class first& $\mathbf{92.42}$ & $\mathbf{76.19}$ \\
    \end{tabular}
    \caption{\textit{Ablation on the hierarchical order.}}
    \label{tab:ablation}
\end{table}
\vspace{-12pt}
\section{Conclusion}
\noindent This paper proposes feature-based losses for hierarchical contrastive learning. The proposed variants, G-HMLC via clustering for small datasets and the end-to-end A-HMLC, operate on the relevant subsets of the embedding space for the respective hierarchical features. This enables fine-grained clustering on all hierarchy levels. As shown in qualitative and quantitative experiments, the superior clustering performance is backed by a $2\%$ improved accuracy on the linear evaluation for balanced and unbalanced hierarchical datasets. A subsequent study on the order of the hierarchy levels demonstrated that feature-based losses effectively utilize all supervisory labels. In this study, we assume that the number of hierarchy levels is known for the employed clustering method. In future work, we will explore the identification of unknown hierarchy levels using a GMM with a Dirichlet process to identify an arbitrary number of clusters in the embedding.
\bibliographystyle{IEEEbib}
\bibliography{refs}

\end{document}